\documentclass{article}

    \usepackage[preprint]{neurips_2024}

\usepackage[utf8]{inputenc} 
\usepackage[T1]{fontenc}    
\usepackage{hyperref}       
\usepackage{url}            
\usepackage{booktabs}       
\usepackage{amsfonts}       
\usepackage{nicefrac}       
\usepackage{microtype}      
\usepackage{xcolor}         
\usepackage{graphicx}
\usepackage{amsmath}
\usepackage{relsize}
\usepackage{subcaption}

\title{Applying Multi-Fidelity Bayesian Optimization in Chemistry: Open Challenges and Major Considerations}

\author{
  Edmund Judge\\
  Department of Chemistry\\
  Imperial College London\\
  United Kingdom\\
  \texttt{ewj23@ic.ac.uk} \\
   \And
Mohammed Azzouzi\\
   Computational Molecular Design Laboratory\\
  EPFL\\
  Switzerland\\
  \texttt{mohammed.azzouzi@epfl.ch} \\
 \And
Austin M. Mroz\\
   Department of Chemistry\\
  Imperial College London\\
  United Kingdom\\
  \texttt{a.mroz@ic.ac.uk} \\
\And
  Antonio Del Rio Chanona\\
  Department of Chemical Engineering\\
  Imperial College London\\
  United Kingdom\\
  \texttt{a.del-rio-chanona@ic.ac.uk} \\
\And
  Kim E. Jelfs
  Department of Chemistry\\
  Imperial College London\\
  United Kingdom\\
  \texttt{k.jelfs@ic.ac.uk} \\
}

\begin{document}
\maketitle

\begin{abstract}
Multi-fidelity Bayesian optimization (MFBO) leverages experimental and/or computational data of varying quality and resource cost to optimize towards desired maxima cost-effectively. This approach is particularly attractive for chemical discovery due to MFBO's ability to integrate diverse data sources. Here, we investigate the application of MFBO to accelerate the identification of promising molecules or materials. We specifically analyze the conditions under which lower-fidelity data can enhance performance compared to single-fidelity problem formulations. We address two key challenges: selecting the optimal acquisition function, understanding the impact of cost, and data fidelity correlation. We then discuss how to assess the effectiveness of MFBO for chemical discovery.\end{abstract}

\section{Introduction}
From new drug molecules, to novel battery technologies, the discovery of new molecules and materials is critical to addressing the global challenges that humanity faces today. Yet, the chemical space that we face is massive; for example, it is estimated that there are $10^{60}$ potential small molecules that could be synthesised. This excludes the number of potential materials that could be made from these molecules, and the optimal experimental conditions for each. This chemical space is far too vast to only search with chemical intuition and the conventional trial-and-improvement process, particularly given laboratory experiments are often slow (up to months) and costly. Often this challenge can be translated into an optimization problem, where we optimize towards improved performance, reaction yield, novelty, \textit{etc.}. Different computational and experimental tools can be used to evaluate the performance of candidate chemical systems, and optimization algorithms then used to select the next best candidates to test. Derivative-free, surrogate-based optimization algorithms are of particular relevance to this task -- specifically, Bayesian optimization (BO).

BO is a resource-aware optimization technique that balances exploration of the parameter space (sampling sparse data regions) with exploitation (sampling regions rich in data). BO has enjoyed a recent resurgence due to advances in computational power that make its implementation more feasible. In practice, implementation is accomplished by using cheap-to-evaluate surrogate models to encode uncertainty about the search-space. With this information, the algorithm will compute, according to user requirements, the next candidate to evaluate in the search space. Within chemistry, BO has been applied to a wide variety of problems, including for antimicrobial polymer design[\citenum{zhai_computational_2023}], electrolyte optimization in zinc-ion batteries[\citenum{gaonkar_multi-objective_2022, jiang_constrained_2022}], and nanoporous materials[\citenum{deshwal_bayesian_2021, gantzler_multi-fidelity_2023}].

While powerful, there are several limitations to applying BO for molecular discovery. First, the datasets used in this domain are often high-dimensional, which causes the surrogate model to suffer from the ``curse of high-dimensionality''[\citenum{malu_bayesian_2021}]. Specifically, the volume of the search space grows exponentially with the number of dimensions used, and therefore the model has to encounter greater uncertainty, which adversely impacts its accuracy. Second, molecular components must be encoded in a machine-readable format. There are several options available, such as SMILES, SELFIES[\citenum{krenn_self-referencing_2020}], molecular graphs, and computer-learned representations, but each has its trade-offs. For example, SMILES are a succinct representation of a molecule using ACSII characters, and are therefore cheap to store and amenable to algorithms from the field of natural language processing; however, critical details of the 3-dimensional configuration of the molecule are lost in this 2-dimensional representation, nor is any feature information communicated - the model having to infer this instead. Furthermore, SMILES strings are non-canonical: this means a molecule can have two distinct SMILES representations that are more varied than two SMILES for totally different molecules[\citenum{wigh_review_2022}]. More geometric and precise representations, such as a chemical table, where the coordinates of each atom in the molecule are recorded in 3-dimensions, also have their drawbacks, since the molecule's invariance to rotations is not intrinsically  captured in the representation. Such inconsistencies can easily confuse a model. One final constraint with BO for molecular discovery is the algorithm's ability to handle mixed-domains, \textit{i.e.} domains containing both categorical and discrete features, which are especially common in chemical design[\citenum{daxberger_mixed-variable_2020, zhang_uncertainty-aware_2022}].

The choice of the kernel, or covariance matrix, is also important. The kernel is a critical component in the optimization process, and can be modified to improve the performance of the surrogate models and BO in general. One approach is to incorporate existing domain knowledge into the kernel to achieve, so-called, physics-informed BO[\citenum{khatamsaz_physics_2023}]. We will not consider this in the investigation.

As chemists, we often have access to several different experiment types that may render related information at different costs[\citenum{schoepfer_cost-informed_2024}]. The plethora of computational techniques already available, of various costs and accuracy, invites the possibility of using the information provided (typically property predications) to supplement the more expensive laboratory experiments. This idea is captured in the notion of a data's fidelity, \textit{i.e.} the accuracy of the data relative to the true quantity. Typically, high-fidelity data is more expensive to obtain, as the cost frequently correlates with the accuracy. Therefore, here we use multi-fidelity BO( MFBO) and consider open challenges associated with the technique[\citenum{fare_multi-fidelity_2022, tran_multi-fidelity_2020, zanjani_foumani_multi-fidelity_2023, palizhati_multi-fidelity_2021, palizhati_agents_2022}]. Specifically, we examine under what circumstances the possession of this lower-fidelity data offers a gain in performance over single-fidelity BO (SFBO). Two challenges are considered, firstly the selection of the acquisition function, and secondly the cost of the low-fidelity data and this data's correlation to the high-fidelity data.

\section{Experimental Methods}
The major difference between the single- and multi-fidelity case is that MFBO enables experiments of varying cost to be performed. Consequently, the aim is no longer to reach the optimum in the fewest number of iterations, but rather to do so while exhausting the smallest budget. Lower-fidelity evaluations usually incur a lower cost, so providing there is some correlation between the low-fidelity and the high-fidelity target, it is possible to optimize the objective function more economically. Here, we briefly outline the general algorithmic framework of MFBO; specifically, we describe the surrogate model and acquisition function(s) that we studied. We encourage readers to engage with more detailed descriptions in the literature for more information[\citenum{gantzler_multi-fidelity_2023, fare_multi-fidelity_2022, tran_multi-fidelity_2020}].

Any model can be used as the surrogate model, providing it encodes some level of uncertainty, which is critical for later stages of the algorithm. Although alternatives do exist, such as Bayesian neural networks[\citenum{springenberg_bayesian_2016, li_study_2024}], here we focused on using Gaussian process (GP) as the surrogate model[\citenum{griffiths_gauche_2023}]. We explore three acquisition functions: i) Multi-Fidelity Maximum Entropy Search (MF-MES)[\citenum{takeno_multi-fidelity_2020}], which measures the gain in mutual information between the candidate element and the maximum function value, ii) Multi-Fidelity Targeted Variance Reduction (MF-TVR)[\citenum{fare_multi-fidelity_2022}], which suggests the element and fidelity that minimize the variance of the model's prediction at the point with the greatest expected improvement after sufficient scaling, and iii) Multi-Fidelity Custom (MF-Custom), a custom acquisition function which normalizes and then combines the outputs of the above two techniques, i.e. \textit{\begin{equation}
    \text{MF-Custom}(X) = \left(\frac{\text{MF-MES}(X)}{||\text{MF-MES}(X)||} + \frac{\text{MF-TVR}(X)} {||\text{MF-TVR}(X)||}\right),
\end{equation}}
\hspace{-0.1cm}where $||\cdot||$ is the Euclidean-norm and $X=[\vec{x_1}, \dots, \vec{x_n}]^T$ is an array of elements, $\vec{x_i}$, from the search-space. We did not include Knowledge Gradient[\citenum{garnett_bayesian_2023}] acquisition function here, as preliminary tests showed the computational cost of the approach was prohibitive. Finally, we compared MFBO to SFBO by running Single-Fidelity Expected-Improvement (SF-EI) alongside each of the above multi-fidelity acquisition functions (see \cite{garnett_bayesian_2023} for a definition).

\section{Results and Discussion}
\subsection{Impact of Problem Formulation on Performance}
We first compare the performance of the different MFBO algorithms with SFBO across four different problems to identify the impact of problem formulation on MFBO performance. Here we consider two synthetic problems and two datasets relevant to chemical discovery.

\textbf{Problem 1}: The first relatively simple problem, involves the synthetic RKHS function[\citenum{assael_iassaelbo-benchmark-rkhs_2019}], with the domain, $[0,1]$, divided into 500 evenly distanced points and low-fidelity data generated by adding Gaussian noise to the high-fidelity evaluations. The function is good for benchmarking BO as it has multiple local maxima to challenge the optimization. We gave an assigned cost of 0.1 to the low-fidelity data relative to the high-fidelity data and there was a correlation of $0.88$ between the two fidelities. 

\textbf{Problem 2}: The second synthetic problem is more challenging, as we consider the 6D negated Hartmann function[\citenum{picheny_benchmark_2013}]. Like before, this is a good benchmarking function, as it has multiple local maxima, as well as the new difficulty of additional dimensions. We produced a dataset using the 6D negated Hartmann function as defined by the BoTorch library[\citenum{balandat_botorch_2020}], with the lower-fidelity data created by adding Gaussian noise to the exact, high-fidelity evaluations (with correlation 0.76) and an assigned relative cost of 0.1. 

\textbf{Problem 3}: For the first chemical discovery problem, we used the dataset employed by Gantzler \textit{et al.} when applying MFBO to locate covalent organic frameworks (COFs), a class of porous materials, with the largest equilibrium absorptive selectivity for xenon and krypton at room temperature (see \cite{gantzler_multi-fidelity_2023} for more details). The lower-fidelity data  was computed to have a correlation 0.97 to the higher-fidelity data, and we assigned a relative cost of 0.2. Each molecule was represented by 4 structural features and 10 compositional features, giving a 14 dimensional vector representation. 

\textbf{Problem 4}: Finally, we looked at a problem using a dataset of organic molecules for organic photovoltaic applications. We built a database of 50,000 molecules using our \textit{stk} supramolecular toolkit software[\citenum{turcani_lukasturcanistk_2024}] that assembles such systems from building blocks. We included common building blocks used in the organic photovoltaic community. The high-fidelity data was property data for these systems computed using the extended-tight binding (xtb) technique[\citenum{bannwarth_extended_2021}], and the lower-fidelity data was the product of a machine-learning model with correlation 0.91, which we assigned a relative cost of 0.1. We represent the molecules as a 72-dimension array of concatenated arrays of building block descriptors calculated using xtb. See Figures~\ref{fig:noise} - \ref{fig:stk_sample} in the appendix for more details of the various datasets, and for a deeper discussion on \texttt{STK{\textunderscore}search}.

Problems 1 and 2 were seeded with 5 initial samples (both high- and low-fidelity evaluations), afforded a budget of 50, and had a domain-size of 500. Problem 3 was seeded with 3 initial samples and allowed to run until the optimum was obtained and had a domain-size of 608. Problem 4 was seeded with 25 initial samples, afforded a budget of 50 and had domain-size of 44928.

\begin{figure}
 \centering
 \includegraphics[width=1\textwidth]{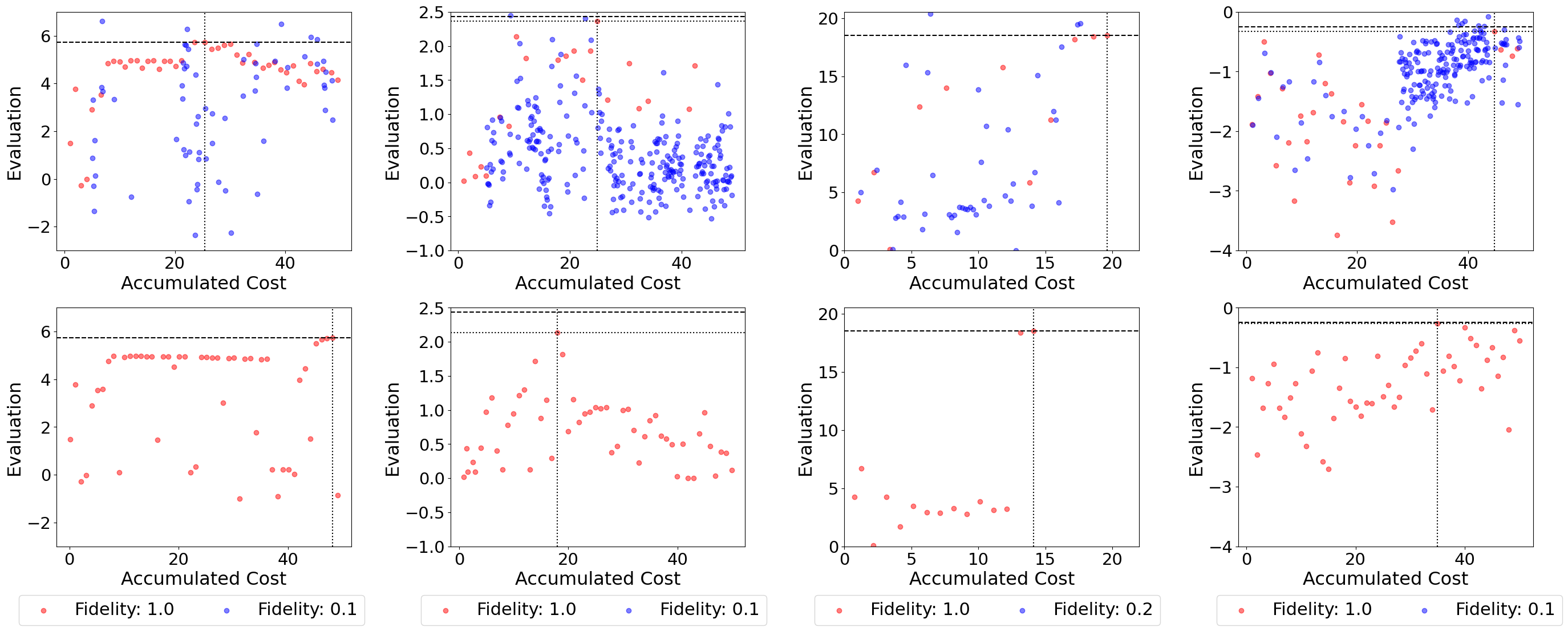}
\caption{The behaviour of the MF-MES (top) and SF-EI (bottom) search-algorithms for a single run optimizing (far-left) Problem 1 (RKHS), (middle-left) Problem 2 (6D negated Hartmann), (middle-right) Problem 3 (COF selectivities) and (far-right) Problem 4 (organic photovoltaic molecules). The different dashed lines denote the domain optimum and the obtained optimum.}
\label{fig:both_single_iterations}
\end{figure}

Figure~\ref{fig:both_single_iterations} illustrates the results of a single run for each of the problems using the MF-MES and SF-EI acquisition functions. MF-MES does not consistently outperform SF-EI, which is surprising as one would expect the presence of the lower-fidelity data to be beneficial. In Problems 1 and 4, there is an improvement in performance with MF-MES obtaining the optimum after a budget of only 25.4 is exhausted (compared to 48.1 for SF-EI), and 9.8 (compared to 35 for SF-EI), respectively. For Problem 4, the domain optimum is not attained in any search. In Problems 2 and 3, MF-MES does not outperform SF-EI, and therefore the presence of the lower-fidelity information does not benefit the optimization process.

\subsection{Fidelity Correlation and Cost}

\begin{figure}
 \centering
 \begin{subfigure}{.49\textwidth}
    \centering
    \includegraphics[width=1\textwidth]{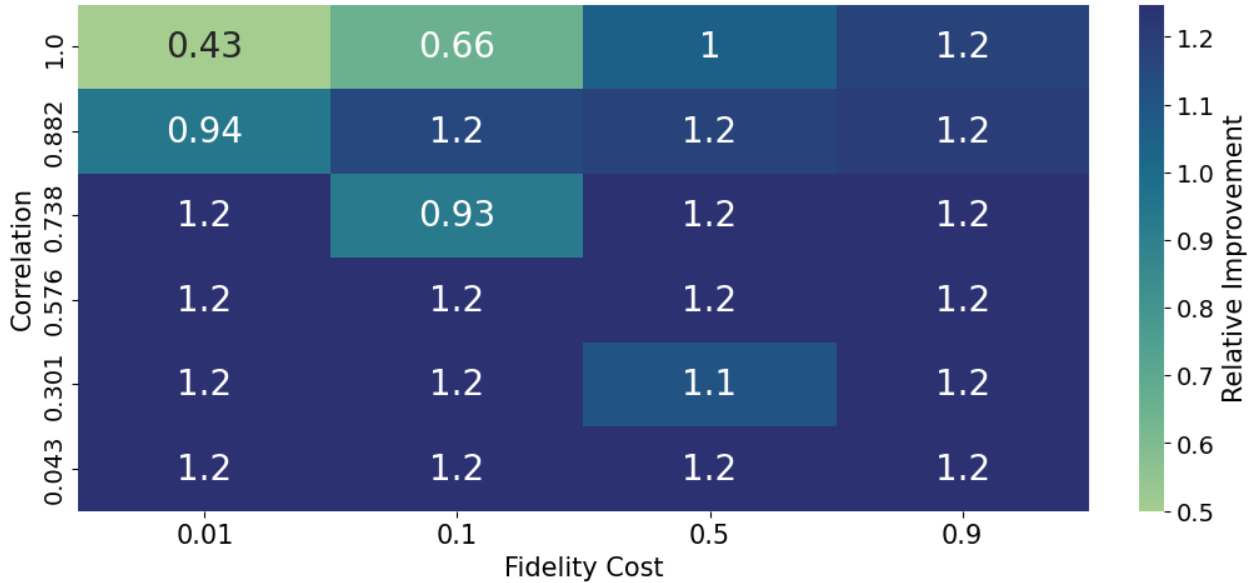}
 \end{subfigure}
 \begin{subfigure}{.49\textwidth}
    \centering
    \includegraphics[width=1\textwidth]{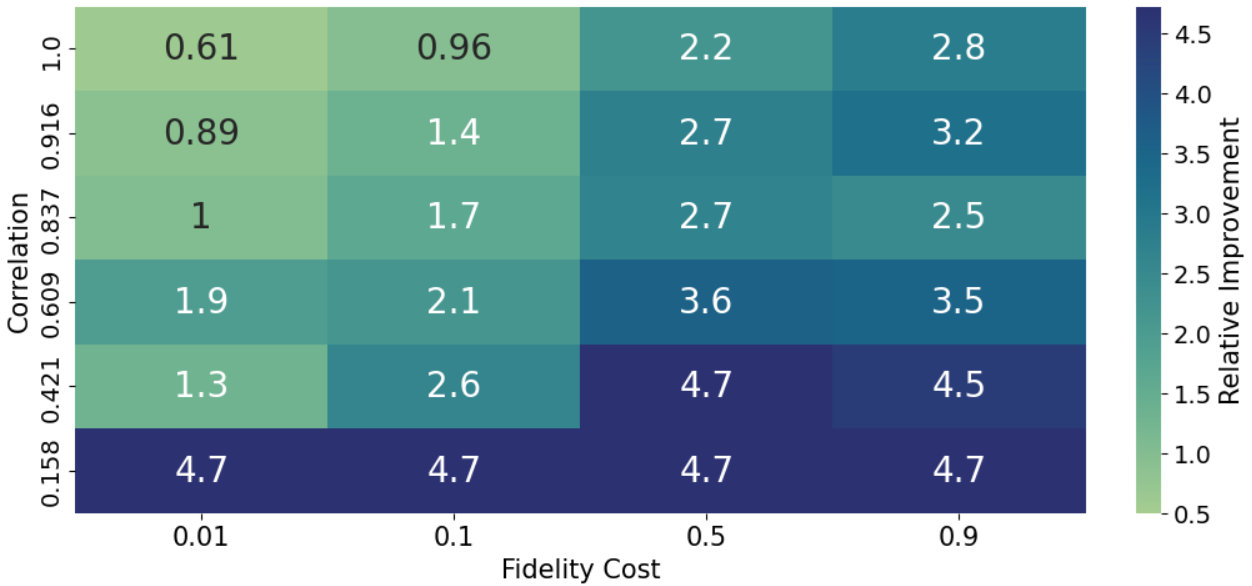}
    \end{subfigure}
\caption{A heatmap for the MF-MES search algorithm showing how the correlation and cost of the low-fidelity data influences the optimization rate for (left) Problem 1 (RKHS) and (right)  Problem 2 (6D negated Hartmann).} 
\label{fig:heatmap_both}
\end{figure}

As shown above, the performance of MFBO depends on the dataset we tackle. The different datasets have different correlations between the high- and low-fidelity data, as well as different problem complexities. To first assess the impact of the cost and the data fidelity correlation, we varied these parameters in the two synthetic problems to see how it affected the relative MFBO performance compared to a single-fidelity search. To do this, new sample-spaces were created for Problems 1 and 2, featuring different quantities of Gaussian noise to alter data correlation. Figure~\ref{fig:heatmap_both} reveals the benefits of high-correlation and low-cost for the MF-MES search-algorithm. Clearly, if the low-fidelity data is highly correlated with the objective function, then the search-algorithm benefits from access to this cheap information. However, what is surprising is how high the correlation and how low the cost of the lower-fidelity data has to be to give the algorithm any sort of advantage over SFBO. Indeed, for Problems 1 and 2, we see only 4 and 3 instances, respectively, where the Relative Improvement is less than 1. We define Relative Improvement as the budget exhausted before obtaining the optimum (averaged over 5 runs, with an allocated budget of 50, and assigned a score of 60 if the optimum was not reached) divided by the budget exhausted by the SF-EI acquisition function (48.1 and 12.7 for Problems 1 and 2, respectively). This is a result seldom discussed explicitly in the literature, but does explain the extreme cost difference between high- and low-fidelity values often observed in well cited papers promoting new acquisition functions[\citenum{gantzler_multi-fidelity_2023, fare_multi-fidelity_2022}]. See Figure~\ref{fig:heatmap_both2} in the appendix for cost-correlation studies for the MF-TVR and MF-Custom formulations.

\subsection{Acquisition Function Selection}
\begin{figure}
 \centering
  \includegraphics[ width=1\textwidth]{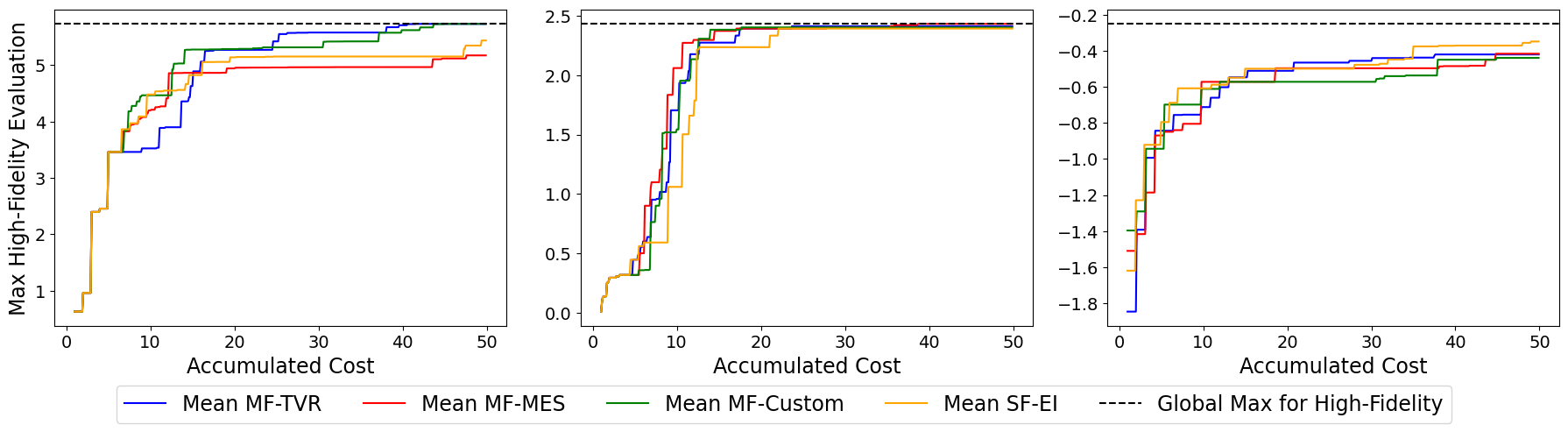}
\caption{Comparison of each of the acquisition functions for Problems (left) 1 (RKHS), (middle) 2 (6D negated Hartmann), (right) 3 (organic photovoltaic molecules).}
\label{fig:batch_plots}
\end{figure}

The performance of MFBO is heavily influenced by the acquisition function used, as this decides which fidelity to evaluate next. Consequently, it is possible to improve the performance by choosing acquisition functions more appropriate for the problem context. Figure~\ref{fig:batch_plots} compares the results across our four Problems when using MF-TVR, MF-MES, MF-Custom and SF-EI as the acquisition function. For Problems 1 and 2, MFBO consistently outperformed SFBO, however SF-EI performed better for the 5 runs for Problem 4. Interestingly, we observed in Figure~\ref{fig:both_single_iterations} how MF-MES outperformed SF-EI with Problem 1 in obtaining the optimum for a single experiment, but see that when averaging over 5 runs, SF-EI outperformed MF-MES. Clearly, MFBO is not always guaranteed to provide a better result. However, the question as to what is even meant by `better' arises, and we discuss what metrics are best to use to measure performance in the next section.

\section{Conclusions and Outlook}

We have explored the use of multi-fidelity Bayesian optimisation (MFBO) for chemical discovery. We found that MFBO outperformed a single fidelity BO (SFBO) in only two of our four problem datasets. These initial results were surprising, as we expected MFBO to, at worst, perform equal to SFBO. We investigated this result firstly by exploring the impact of the low-fidelity data cost and its correlation to the high-fidelity data, finding that MFBO only consistently outperforms SFBO in cases where the cost of the low-fidelity data is below 10 to 100 times the high-fidelity one and with a high correlation above 0.9. This requirement limits the application of MFBO for chemical discovery, and should be addressed before its widespread implementation. Secondly, we explored the use of different acquisition functions. We considered three different acquisition functions and found that their performance was problem specific. This suggests the acquisition function should be carefully adopted for each case explored. 

Simply measuring how efficiently the optimization process locates the maximum is not always a representative metric of the algorithm's performance, since in key applications, such as molecular discovery, the objective is not only to find the global maximum, but to seek multiple high-performing candidates. The metrics of {\it{instantaneous regret}} and {\it{cumulative regret}} are useful here, these can be used to measure how consistently high-performing selected candidates are, even once the optimum has been discovered. However, such metrics are not designed for the multi-fidelity setting, as regret simply takes into account an evaluation's distance from the high-fidelity optimum. The application to a low-fidelity evaluation is thus less clear. Therefore, we suggest a custom metric of `cumulative regret per high-fidelity evaluation' ($CRHF$), where:
\begin{equation}
    CRHF = \mathlarger{\sum_i}\frac{\rho(i)}{\sum_{j=0}^i HF(j)}
\end{equation}
so that the budget is divided into intervals $i$, $\rho(i)$ takes the most recent high-fidelity evaluation at $i$, and $\sum_{j=0}^{i} HF(j)$ is the number of high-fidelity evaluations up to $i$. When a low-fidelity value is selected, the metric takes the previous high-fidelity evaluation; however, to penalize a technique that simply takes low-fidelity evaluations repeatedly, the value is divided by how many high-fidelity values have been evaluated so far. See Figures~\ref{fig:rkhs_new_metric} and \ref{fig:stk_new_metric} in the appendix for more results.

\section*{Code and Data Availability}
The code for applying the search-algorithms to the RKHS and Hartmann functions, as well as to the COF dataset created by Gantzler et al., is available at \url{github.com/kernelCruncher/MFBO}. The code used for Problem 4 is available at \url{github.com/mohammedazzouzi15/STK_search/tree/master}.

\bibliography{MScDissertation} 

\providecommand*{\mcitethebibliography}{\thebibliography}
\csname @ifundefined\endcsname{endmcitethebibliography}
{\let\endmcitethebibliography\endthebibliography}{}
\begin{mcitethebibliography}{31}
\providecommand*{\natexlab}[1]{#1}
\providecommand*{\mciteSetBstSublistMode}[1]{}
\providecommand*{\mciteSetBstMaxWidthForm}[2]{}
\providecommand*{\mciteBstWouldAddEndPuncttrue}
  {\def\EndOfBibitem{\unskip.}}
\providecommand*{\mciteBstWouldAddEndPunctfalse}
  {\let\EndOfBibitem\relax}
\providecommand*{\mciteSetBstMidEndSepPunct}[3]{}
\providecommand*{\mciteSetBstSublistLabelBeginEnd}[3]{}
\providecommand*{\EndOfBibitem}{}
\mciteSetBstSublistMode{f}
\mciteSetBstMaxWidthForm{subitem}
{(\emph{\alph{mcitesubitemcount}})}
\mciteSetBstSublistLabelBeginEnd{\mcitemaxwidthsubitemform\space}
{\relax}{\relax}

\bibitem[Zhai and Yeo(2023)]{zhai_computational_2023}
H.~Zhai and J.~Yeo, \emph{ACS Biomaterials Science \& Engineering}, 2023, \textbf{9}, 269--279\relax
\mciteBstWouldAddEndPuncttrue
\mciteSetBstMidEndSepPunct{\mcitedefaultmidpunct}
{\mcitedefaultendpunct}{\mcitedefaultseppunct}\relax
\EndOfBibitem
\bibitem[Gaonkar \emph{et~al.}(2022)Gaonkar, Valladares, Tovar, Zhu, and El-Mounayri]{gaonkar_multi-objective_2022}
A.~Gaonkar, H.~Valladares, A.~Tovar, L.~Zhu and H.~El-Mounayri, \emph{Electronic Materials}, 2022, \textbf{3}, 201--217\relax
\mciteBstWouldAddEndPuncttrue
\mciteSetBstMidEndSepPunct{\mcitedefaultmidpunct}
{\mcitedefaultendpunct}{\mcitedefaultseppunct}\relax
\EndOfBibitem
\bibitem[Jiang and Wang(2022)]{jiang_constrained_2022}
B.~Jiang and X.~Wang, \emph{IEEE Control Systems Letters}, 2022, \textbf{6}, 1682--1687\relax
\mciteBstWouldAddEndPuncttrue
\mciteSetBstMidEndSepPunct{\mcitedefaultmidpunct}
{\mcitedefaultendpunct}{\mcitedefaultseppunct}\relax
\EndOfBibitem
\bibitem[Deshwal \emph{et~al.}(2021)Deshwal, Simon, and Doppa]{deshwal_bayesian_2021}
A.~Deshwal, C.~M. Simon and J.~R. Doppa, \emph{Molecular Systems Design \& Engineering}, 2021, \textbf{6}, 1066--1086\relax
\mciteBstWouldAddEndPuncttrue
\mciteSetBstMidEndSepPunct{\mcitedefaultmidpunct}
{\mcitedefaultendpunct}{\mcitedefaultseppunct}\relax
\EndOfBibitem
\bibitem[Gantzler \emph{et~al.}(2023)Gantzler, Deshwal, Doppa, and Simon]{gantzler_multi-fidelity_2023}
N.~Gantzler, A.~Deshwal, J.~R. Doppa and C.~M. Simon, \emph{Digital Discovery}, 2023, \textbf{2}, 1937--1956\relax
\mciteBstWouldAddEndPuncttrue
\mciteSetBstMidEndSepPunct{\mcitedefaultmidpunct}
{\mcitedefaultendpunct}{\mcitedefaultseppunct}\relax
\EndOfBibitem
\bibitem[Malu \emph{et~al.}(2021)Malu, Dasarathy, and Spanias]{malu_bayesian_2021}
M.~Malu, G.~Dasarathy and A.~Spanias, 2021 12th {International} {Conference} on {Information}, {Intelligence}, {Systems} \& {Applications} ({IISA}), Chania Crete, Greece, 2021, pp. 1--8\relax
\mciteBstWouldAddEndPuncttrue
\mciteSetBstMidEndSepPunct{\mcitedefaultmidpunct}
{\mcitedefaultendpunct}{\mcitedefaultseppunct}\relax
\EndOfBibitem
\bibitem[Krenn \emph{et~al.}(2020)Krenn, Häse, Nigam, Friederich, and Aspuru-Guzik]{krenn_self-referencing_2020}
M.~Krenn, F.~Häse, A.~Nigam, P.~Friederich and A.~Aspuru-Guzik, \emph{Machine Learning: Science and Technology}, 2020, \textbf{1}, 045024\relax
\mciteBstWouldAddEndPuncttrue
\mciteSetBstMidEndSepPunct{\mcitedefaultmidpunct}
{\mcitedefaultendpunct}{\mcitedefaultseppunct}\relax
\EndOfBibitem
\bibitem[Wigh \emph{et~al.}(2022)Wigh, Goodman, and Lapkin]{wigh_review_2022}
D.~S. Wigh, J.~M. Goodman and A.~A. Lapkin, \emph{WIREs Computational Molecular Science}, 2022, \textbf{12}, e1603\relax
\mciteBstWouldAddEndPuncttrue
\mciteSetBstMidEndSepPunct{\mcitedefaultmidpunct}
{\mcitedefaultendpunct}{\mcitedefaultseppunct}\relax
\EndOfBibitem
\bibitem[Daxberger \emph{et~al.}(2020)Daxberger, Makarova, Turchetta, and Krause]{daxberger_mixed-variable_2020}
E.~Daxberger, A.~Makarova, M.~Turchetta and A.~Krause, Proceedings of the {Twenty}-{Ninth} {International} {Joint} {Conference} on {Artificial} {Intelligence}, Yokohama, Japan, 2020, pp. 2633--2639\relax
\mciteBstWouldAddEndPuncttrue
\mciteSetBstMidEndSepPunct{\mcitedefaultmidpunct}
{\mcitedefaultendpunct}{\mcitedefaultseppunct}\relax
\EndOfBibitem
\bibitem[Zhang \emph{et~al.}(2022)Zhang, Chen, Iyer, Apley, and Chen]{zhang_uncertainty-aware_2022}
H.~Zhang, W.~W. Chen, A.~Iyer, D.~W. Apley and W.~Chen, \emph{Uncertainty-{Aware} {Mixed}-{Variable} {Machine} {Learning} for {Materials} {Design}}, 2022, \url{https://www.researchsquare.com/article/rs-1987975/v1}\relax
\mciteBstWouldAddEndPuncttrue
\mciteSetBstMidEndSepPunct{\mcitedefaultmidpunct}
{\mcitedefaultendpunct}{\mcitedefaultseppunct}\relax
\EndOfBibitem
\bibitem[Khatamsaz \emph{et~al.}(2023)Khatamsaz, Neuberger, Roy, Zadeh, Otis, and Arróyave]{khatamsaz_physics_2023}
D.~Khatamsaz, R.~Neuberger, A.~M. Roy, S.~H. Zadeh, R.~Otis and R.~Arróyave, \emph{npj Computational Materials}, 2023, \textbf{9}, 221\relax
\mciteBstWouldAddEndPuncttrue
\mciteSetBstMidEndSepPunct{\mcitedefaultmidpunct}
{\mcitedefaultendpunct}{\mcitedefaultseppunct}\relax
\EndOfBibitem
\bibitem[Schoepfer \emph{et~al.}(2024)Schoepfer, Weinreich, Laplaza, Waser, and Corminboeuf]{schoepfer_cost-informed_2024}
A.~Schoepfer, J.~Weinreich, R.~Laplaza, J.~Waser and C.~Corminboeuf, \emph{Cost-{Informed} {Bayesian} {Reaction} {Optimization}}, 2024, \url{https://chemrxiv.org/engage/chemrxiv/article-details/66220e8a21291e5d1d27408d}\relax
\mciteBstWouldAddEndPuncttrue
\mciteSetBstMidEndSepPunct{\mcitedefaultmidpunct}
{\mcitedefaultendpunct}{\mcitedefaultseppunct}\relax
\EndOfBibitem
\bibitem[Fare \emph{et~al.}(2022)Fare, Fenner, Benatan, Varsi, and Pyzer-Knapp]{fare_multi-fidelity_2022}
C.~Fare, P.~Fenner, M.~Benatan, A.~Varsi and E.~O. Pyzer-Knapp, \emph{npj Computational Materials}, 2022, \textbf{8}, 257\relax
\mciteBstWouldAddEndPuncttrue
\mciteSetBstMidEndSepPunct{\mcitedefaultmidpunct}
{\mcitedefaultendpunct}{\mcitedefaultseppunct}\relax
\EndOfBibitem
\bibitem[Tran \emph{et~al.}(2020)Tran, Tranchida, Wildey, and Thompson]{tran_multi-fidelity_2020}
A.~Tran, J.~Tranchida, T.~Wildey and A.~P. Thompson, \emph{The Journal of Chemical Physics}, 2020, \textbf{153}, 074705\relax
\mciteBstWouldAddEndPuncttrue
\mciteSetBstMidEndSepPunct{\mcitedefaultmidpunct}
{\mcitedefaultendpunct}{\mcitedefaultseppunct}\relax
\EndOfBibitem
\bibitem[Zanjani~Foumani \emph{et~al.}(2023)Zanjani~Foumani, Shishehbor, Yousefpour, and Bostanabad]{zanjani_foumani_multi-fidelity_2023}
Z.~Zanjani~Foumani, M.~Shishehbor, A.~Yousefpour and R.~Bostanabad, \emph{Computer Methods in Applied Mechanics and Engineering}, 2023, \textbf{407}, 115937\relax
\mciteBstWouldAddEndPuncttrue
\mciteSetBstMidEndSepPunct{\mcitedefaultmidpunct}
{\mcitedefaultendpunct}{\mcitedefaultseppunct}\relax
\EndOfBibitem
\bibitem[Palizhati \emph{et~al.}(2021)Palizhati, Aykol, Suram, Hummelshøj, and Montoya]{palizhati_multi-fidelity_2021}
A.~Palizhati, M.~Aykol, S.~Suram, J.~S. Hummelshøj and J.~H. Montoya, \emph{Multi-fidelity {Sequential} {Learning} for {Accelerated} {Materials} {Discovery}}, 2021, \url{https://chemrxiv.org/engage/chemrxiv/article-details/60c756c60f50dbb7f939813f}\relax
\mciteBstWouldAddEndPuncttrue
\mciteSetBstMidEndSepPunct{\mcitedefaultmidpunct}
{\mcitedefaultendpunct}{\mcitedefaultseppunct}\relax
\EndOfBibitem
\bibitem[Palizhati \emph{et~al.}(2022)Palizhati, Torrisi, Aykol, Suram, Hummelshøj, and Montoya]{palizhati_agents_2022}
A.~Palizhati, S.~B. Torrisi, M.~Aykol, S.~K. Suram, J.~S. Hummelshøj and J.~H. Montoya, \emph{Scientific Reports}, 2022, \textbf{12}, 4694\relax
\mciteBstWouldAddEndPuncttrue
\mciteSetBstMidEndSepPunct{\mcitedefaultmidpunct}
{\mcitedefaultendpunct}{\mcitedefaultseppunct}\relax
\EndOfBibitem
\bibitem[Springenberg \emph{et~al.}(2016)Springenberg, Klein, Falkner, and Hutter]{springenberg_bayesian_2016}
J.~T. Springenberg, A.~Klein, S.~Falkner and F.~Hutter, Advances in {Neural} {Information} {Processing} {Systems}, 2016\relax
\mciteBstWouldAddEndPuncttrue
\mciteSetBstMidEndSepPunct{\mcitedefaultmidpunct}
{\mcitedefaultendpunct}{\mcitedefaultseppunct}\relax
\EndOfBibitem
\bibitem[Li \emph{et~al.}(2024)Li, Rudner, and Wilson]{li_study_2024}
Y.~L. Li, T.~G.~J. Rudner and A.~G. Wilson, \emph{A {Study} of {Bayesian} {Neural} {Network} {Surrogates} for {Bayesian} {Optimization}}, 2024, \url{http://arxiv.org/abs/2305.20028}, arXiv:2305.20028 [cs, stat]\relax
\mciteBstWouldAddEndPuncttrue
\mciteSetBstMidEndSepPunct{\mcitedefaultmidpunct}
{\mcitedefaultendpunct}{\mcitedefaultseppunct}\relax
\EndOfBibitem
\bibitem[Griffiths \emph{et~al.}(2023)Griffiths, Klarner, Moss, Ravuri, Truong, Stanton, Tom, Rankovic, Du, Jamasb, Deshwal, Schwartz, Tripp, Kell, Frieder, Bourached, Chan, Moss, Guo, Durholt, Chaurasia, Strieth-Kalthoff, Lee, Cheng, Aspuru-Guzik, Schwaller, and Tang]{griffiths_gauche_2023}
R.-R. Griffiths, L.~Klarner, H.~B. Moss, A.~Ravuri, S.~Truong, S.~Stanton, G.~Tom, B.~Rankovic, Y.~Du, A.~Jamasb, A.~Deshwal, J.~Schwartz, A.~Tripp, G.~Kell, S.~Frieder, A.~Bourached, A.~Chan, J.~Moss, C.~Guo, J.~Durholt, S.~Chaurasia, F.~Strieth-Kalthoff, A.~A. Lee, B.~Cheng, A.~Aspuru-Guzik, P.~Schwaller and J.~Tang, \emph{{GAUCHE}: {A} {Library} for {Gaussian} {Processes} in {Chemistry}}, 2023, \url{http://arxiv.org/abs/2212.04450}, arXiv:2212.04450 [cond-mat, physics:physics]\relax
\mciteBstWouldAddEndPuncttrue
\mciteSetBstMidEndSepPunct{\mcitedefaultmidpunct}
{\mcitedefaultendpunct}{\mcitedefaultseppunct}\relax
\EndOfBibitem
\bibitem[Takeno \emph{et~al.}(2020)Takeno, Fukuoka, Tsukada, Koyama, Shiga, Takeuchi, and Karasuyama]{takeno_multi-fidelity_2020}
S.~Takeno, H.~Fukuoka, Y.~Tsukada, T.~Koyama, M.~Shiga, I.~Takeuchi and M.~Karasuyama, \emph{Multi-fidelity {Bayesian} {Optimization} with {Max}-value {Entropy} {Search} and its parallelization}, 2020, \url{http://arxiv.org/abs/1901.08275}, arXiv:1901.08275 [cs, stat]\relax
\mciteBstWouldAddEndPuncttrue
\mciteSetBstMidEndSepPunct{\mcitedefaultmidpunct}
{\mcitedefaultendpunct}{\mcitedefaultseppunct}\relax
\EndOfBibitem
\bibitem[Garnett(2023)]{garnett_bayesian_2023}
R.~Garnett, \emph{Bayesian {Optimization}}, Cambridge University Press, Cambridge, 2023\relax
\mciteBstWouldAddEndPuncttrue
\mciteSetBstMidEndSepPunct{\mcitedefaultmidpunct}
{\mcitedefaultendpunct}{\mcitedefaultseppunct}\relax
\EndOfBibitem
\bibitem[Assael(2019)]{assael_iassaelbo-benchmark-rkhs_2019}
Y.~Assael, \emph{iassael/bo-benchmark-rkhs}, 2019, \url{https://github.com/iassael/bo-benchmark-rkhs}, original-date: 2014-10-18T14:58:36Z\relax
\mciteBstWouldAddEndPuncttrue
\mciteSetBstMidEndSepPunct{\mcitedefaultmidpunct}
{\mcitedefaultendpunct}{\mcitedefaultseppunct}\relax
\EndOfBibitem
\bibitem[Picheny \emph{et~al.}(2013)Picheny, Wagner, and Ginsbourger]{picheny_benchmark_2013}
V.~Picheny, T.~Wagner and D.~Ginsbourger, \emph{Structural and Multidisciplinary Optimization}, 2013, \textbf{48}, 607--626\relax
\mciteBstWouldAddEndPuncttrue
\mciteSetBstMidEndSepPunct{\mcitedefaultmidpunct}
{\mcitedefaultendpunct}{\mcitedefaultseppunct}\relax
\EndOfBibitem
\bibitem[Balandat \emph{et~al.}(2020)Balandat, Karrer, Jiang, Daulton, Letham, Wilson, and Bakshy]{balandat_botorch_2020}
M.~Balandat, B.~Karrer, D.~R. Jiang, S.~Daulton, B.~Letham, A.~G. Wilson and E.~Bakshy, \emph{{BoTorch}: {A} {Framework} for {Efficient} {Monte}-{Carlo} {Bayesian} {Optimization}}, 2020, \url{http://arxiv.org/abs/1910.06403}, arXiv:1910.06403 [cs, math, stat]\relax
\mciteBstWouldAddEndPuncttrue
\mciteSetBstMidEndSepPunct{\mcitedefaultmidpunct}
{\mcitedefaultendpunct}{\mcitedefaultseppunct}\relax
\EndOfBibitem
\bibitem[Turcani(2024)]{turcani_lukasturcanistk_2024}
L.~Turcani, \emph{lukasturcani/stk}, 2024, \url{https://github.com/lukasturcani/stk}, original-date: 2018-03-18T20:57:46Z\relax
\mciteBstWouldAddEndPuncttrue
\mciteSetBstMidEndSepPunct{\mcitedefaultmidpunct}
{\mcitedefaultendpunct}{\mcitedefaultseppunct}\relax
\EndOfBibitem
\bibitem[Bannwarth \emph{et~al.}(2021)Bannwarth, Caldeweyher, Ehlert, Hansen, Pracht, Seibert, Spicher, and Grimme]{bannwarth_extended_2021}
C.~Bannwarth, E.~Caldeweyher, S.~Ehlert, A.~Hansen, P.~Pracht, J.~Seibert, S.~Spicher and S.~Grimme, \emph{WIREs Computational Molecular Science}, 2021, \textbf{11}, e1493\relax
\mciteBstWouldAddEndPuncttrue
\mciteSetBstMidEndSepPunct{\mcitedefaultmidpunct}
{\mcitedefaultendpunct}{\mcitedefaultseppunct}\relax
\EndOfBibitem
\bibitem[Riniker and Landrum(2015)]{Riniker2015}
S.~Riniker and G.~A. Landrum, \emph{Journal of Chemical Information and Modeling}, 2015, \textbf{55}, 2562--2574\relax
\mciteBstWouldAddEndPuncttrue
\mciteSetBstMidEndSepPunct{\mcitedefaultmidpunct}
{\mcitedefaultendpunct}{\mcitedefaultseppunct}\relax
\EndOfBibitem
\bibitem[Grimme(2019)]{grimme2019}
S.~Grimme, \emph{J{.} Chem{.} Theory Comput{.}}, 2019, \textbf{15}, 2847--2862\relax
\mciteBstWouldAddEndPuncttrue
\mciteSetBstMidEndSepPunct{\mcitedefaultmidpunct}
{\mcitedefaultendpunct}{\mcitedefaultseppunct}\relax
\EndOfBibitem
\bibitem[Grimme and Bannwarth(2016)]{grimme2016}
S.~Grimme and C.~Bannwarth, \emph{J{.} Chem{.} Phys{.}}, 2016, \textbf{145}, 054103\relax
\mciteBstWouldAddEndPuncttrue
\mciteSetBstMidEndSepPunct{\mcitedefaultmidpunct}
{\mcitedefaultendpunct}{\mcitedefaultseppunct}\relax
\EndOfBibitem
\bibitem[Schütt \emph{et~al.}(2018)Schütt, Sauceda, Kindermans, Tkatchenko, and Müller]{Schütt2018}
K.~T. Schütt, H.~E. Sauceda, P.-J. Kindermans, A.~Tkatchenko and K.-R. Müller, \emph{The Journal of Chemical Physics}, 2018, \textbf{148}, 241722\relax
\mciteBstWouldAddEndPuncttrue
\mciteSetBstMidEndSepPunct{\mcitedefaultmidpunct}
{\mcitedefaultendpunct}{\mcitedefaultseppunct}\relax
\EndOfBibitem
\end{mcitethebibliography}
\bibliographystyle{rsc}
\newpage

\appendix
\section{Appendix}

\subsection{Dataset Distributions}

\begin{figure}[h!]
    \centering
    \includegraphics[width=0.8\textwidth]{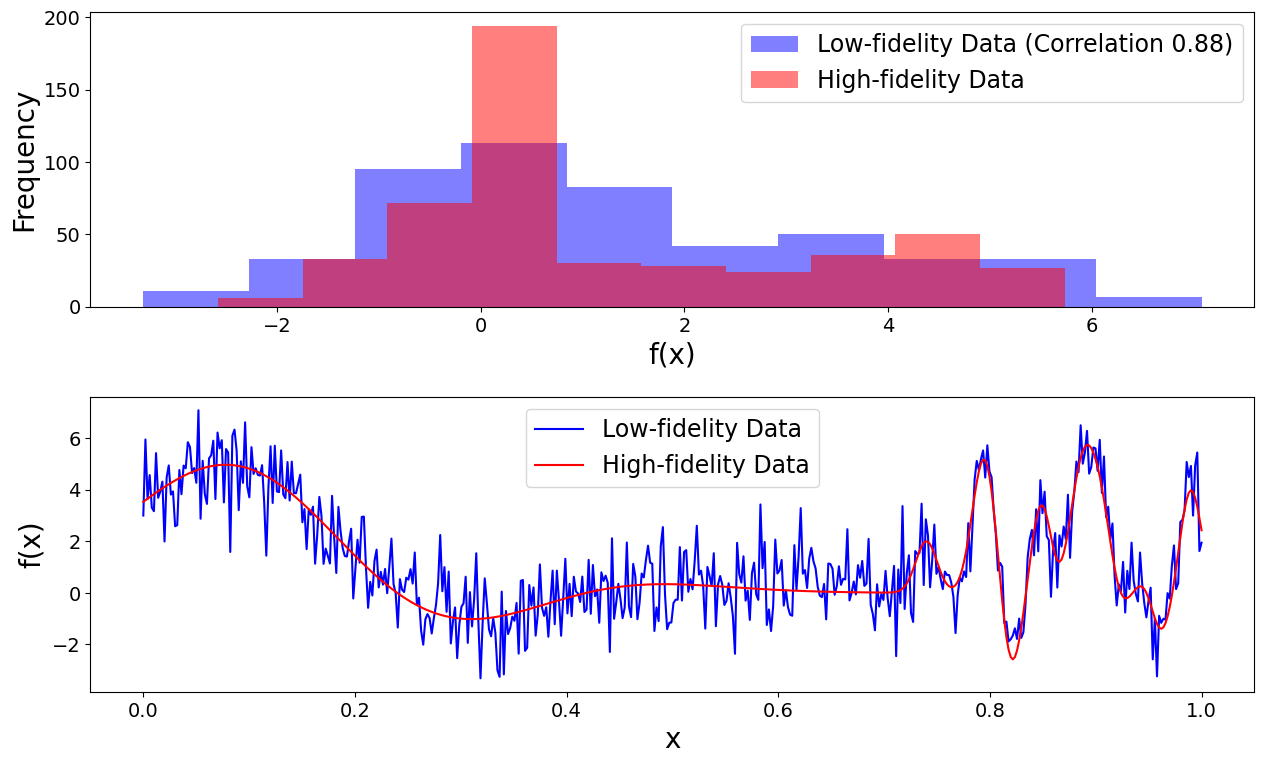}
    \caption{High- and low-fidelity data for the RKHS function and their distributions (Problem 1).} 
    \label{fig:noise}
\end{figure}

\begin{figure}[h!]
\centering
\includegraphics[width=0.8\textwidth]{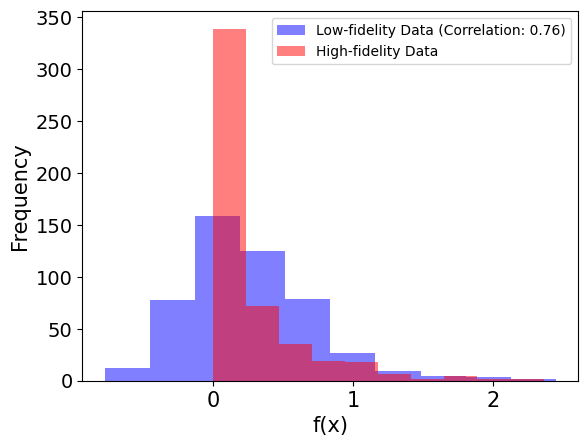}
\caption{Histogram of 6D Hartmann evaluations (Problem 2).} 
\label{fig:hartmann_sample}
\end{figure}

\begin{figure}[h!]
\centering
\includegraphics[width=0.8\textwidth]{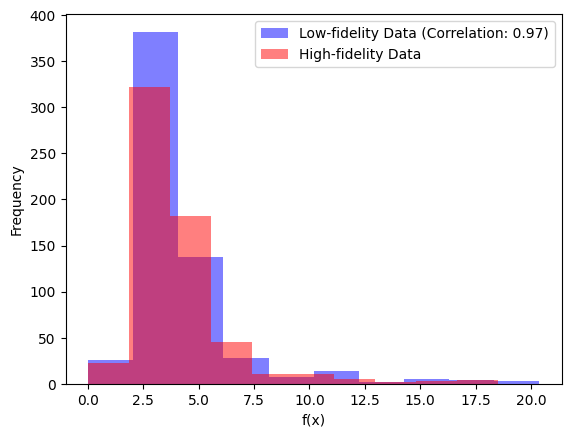}
\caption{Histogram of COF selectivities (Problem 3).} 
\label{fig:CoreySimonsDomain}
\end{figure}

\clearpage

\subsection{\texttt{STK{\textunderscore}search} dataset}

The molecules in \texttt{STK{\textunderscore}search} are composed of 6 building-blocks, or units, each of which has 306 different candidate structures. This creates a combinatorial space of $306^6=8\times 10^{14}$ possible oligomers to be explored. From this space we only consider 45,000 molecules with 30,000 randomly selected and 15,000 generated through trying different optimisation algorithms.  Each unit is fed into a deep learning model that embeds its features, such as HOMO, LUMO, excited state energy, etc, into 12-dimensional vector space, thus causing the whole molecule to be represented by a 72-dimensional vector. 
For the \texttt{STK{\textunderscore}search}, the BO proceeds similarly to what is described in the main text, namely the GP is seeded with an initial sample, which an acquisition function then uses to determine the most valuable elements to evaluate next. However, due to the size of the space it no longer becomes computationally feasible to apply the acquisition function to every element, and instead, an evolutionary algorithmic approach is applied that combines the building-blocks of the strongest candidates, so far, whilst also injecting randomness, via mutations, to ensure a wide, albeit incomplete, search of the space. As the name suggests, evolutionary algorithms are inspired by biology, specifically natural selection, where the unit-like structure resembles a sequence of genes competing for dominance.

We optimize the molecules to align with the specific properties of a molecular acceptor, namely ionisation potential and complimentary absorption. The property of the molecule to be predicted, therefore, is the combined-target function, $F_{comb}$, defined as 
\begin{equation}
F_{comb} = -|E_{S1} - 3| - |IP - 5.5| + \log(f_{osc,S1}),
\end{equation}
where $IP$ is the ionization potential, $E_{S1}$ is the first excited state energy, and $f_{osc,S1}$ is the oscillator strength of the first excited state. To calculate these properties,  we used stk to generate initial geometries for the molecules, followed by the use of the Experimental-Torsion basic Knowledge Distance Geometry (ETKDG) approach in stk/RDkit to generate a first geometry.[\citenum{Riniker2015}] Then, we optimised the geometry of the lowest energy conformer found using GFN2-XTB [\citenum{grimme2019}] and calculated the ionisation potential and electron affinity using the IPEA option in XTB. The optical properties of the molecules were calculated using sTDA-XTB. [\citenum{grimme2016}]
The lower-fidelity data is produced by the predictions of a machine-learning model trained on the same dataset. Here we used Schnet as the surrogate model[\citenum{Schütt2018}]. The low- and high-fidelity data are available at \url{github.com/kernelCruncher/MFBO}.
We assigned an initial cost of 0.1, although there is some discussion surrounding the significance of this cost-value and its impact to the MFBO process. Indeed, we posit that the selection of fidelity values should be an ongoing area of research, with design-principles yet unknown and problem-specific.

\begin{figure}[h!]
\centering
\includegraphics[width=0.8\textwidth]{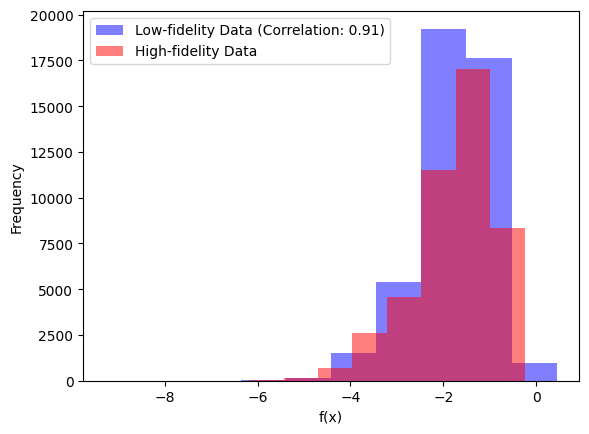}
\caption{Histogram of combined-target evaluations for the organic photovoltaic molecules (Problem 4).} 
\label{fig:stk_sample}
\end{figure}

\clearpage
\subsection{Fidelity Cost and Correlation Heatmaps}

\begin{figure}[h!]
 \centering
 \begin{subfigure}{.49\textwidth}
    \centering
    \includegraphics[width=1\textwidth]{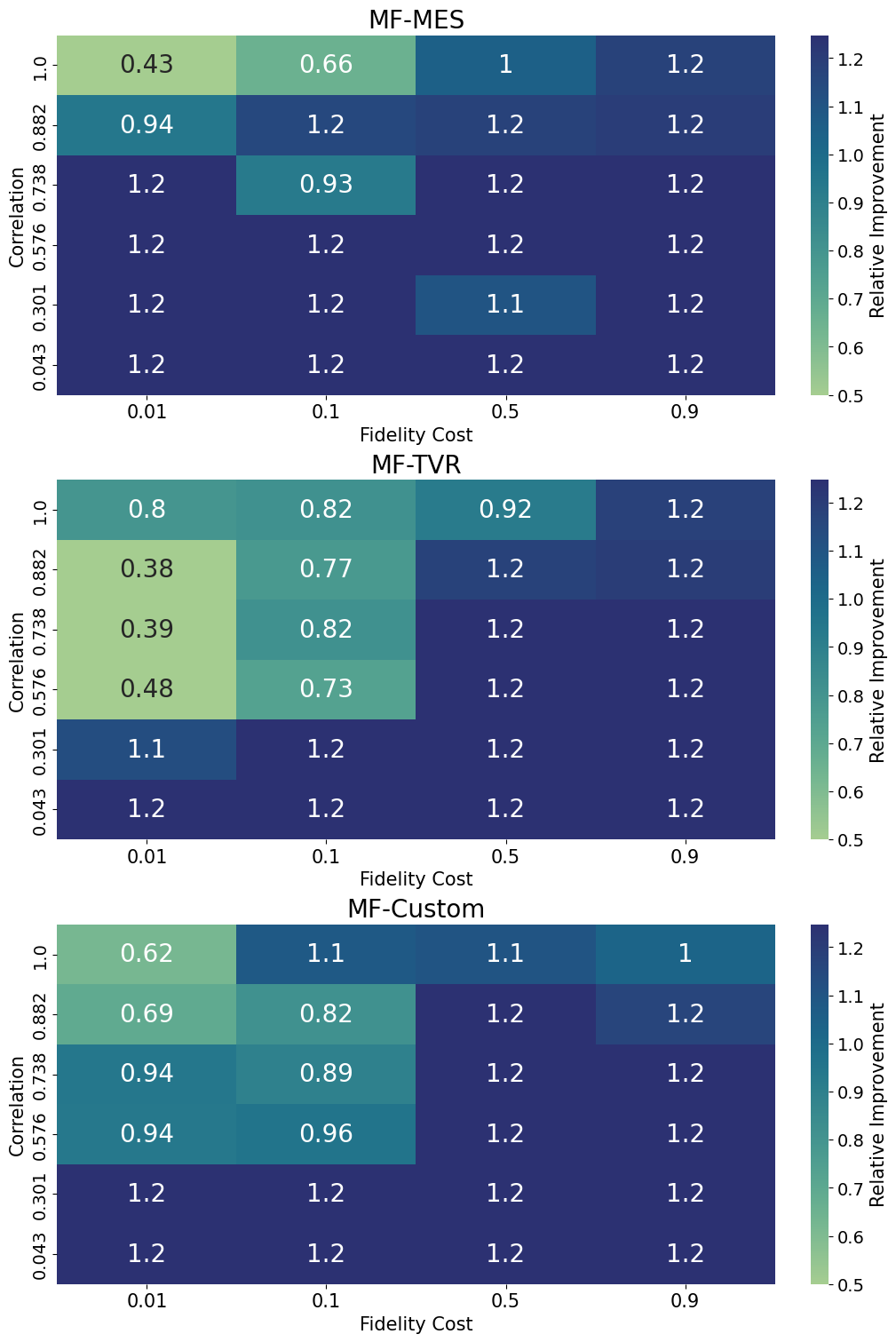}
        \caption{RKHS}
 \end{subfigure}
 \begin{subfigure}{.49\textwidth}
    \centering
    \includegraphics[width=1\textwidth]{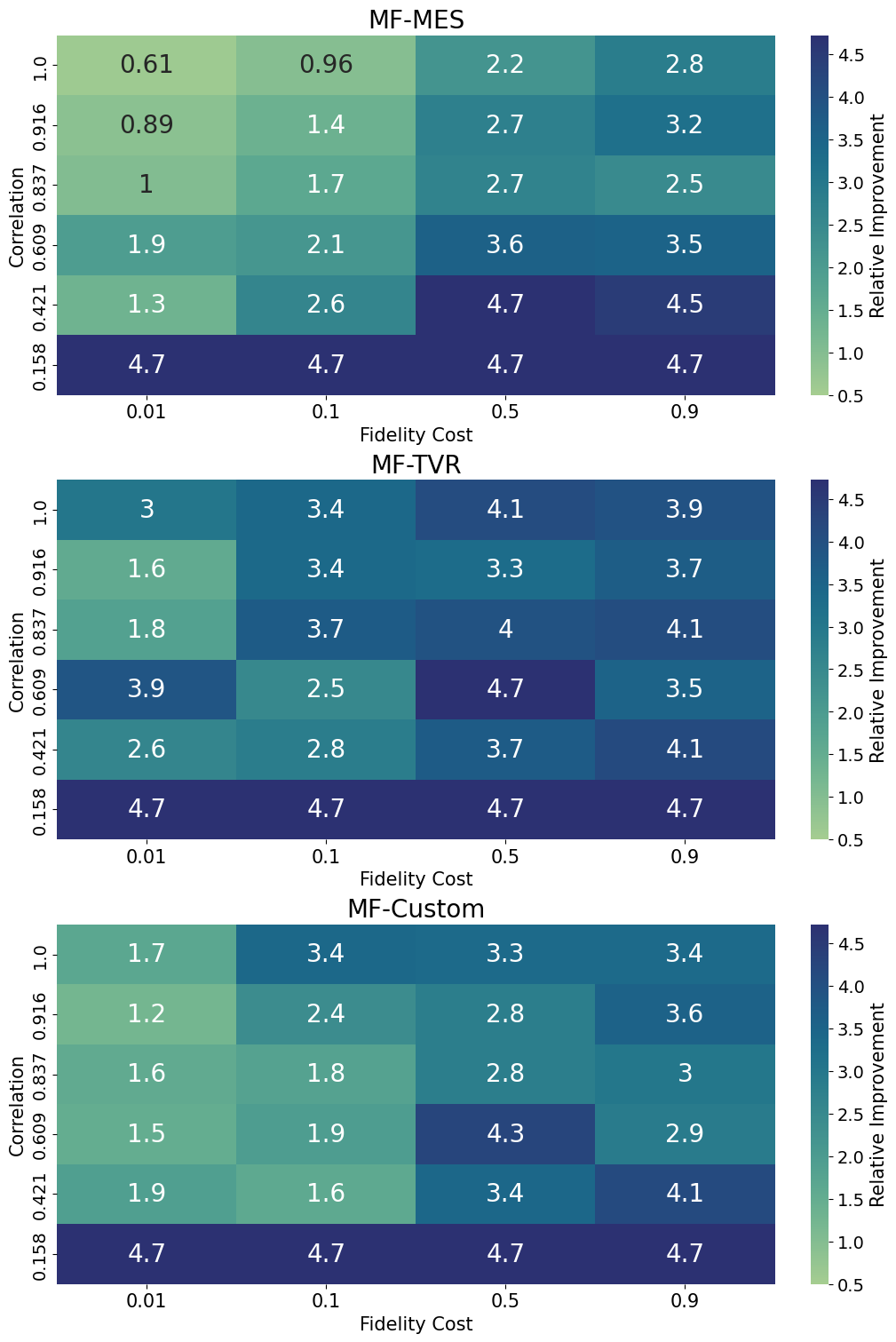}
    \caption{6D negated Hartmann}
    \end{subfigure}
\caption{A heatmap for the MF-MES, MF-TVR and MF-Custom acquisition functions illustrating how the correlation and cost of of the low-fidelity data influences the rate of optimization for (a) RKHS (Problem 1) and (b)  6D negated Hartmann (Problem 2).} 
\label{fig:heatmap_both2}
\end{figure}
\clearpage
\subsection{Custom Metrics}

\begin{figure}[h!]
    \centering
    \includegraphics[width=1\textwidth]{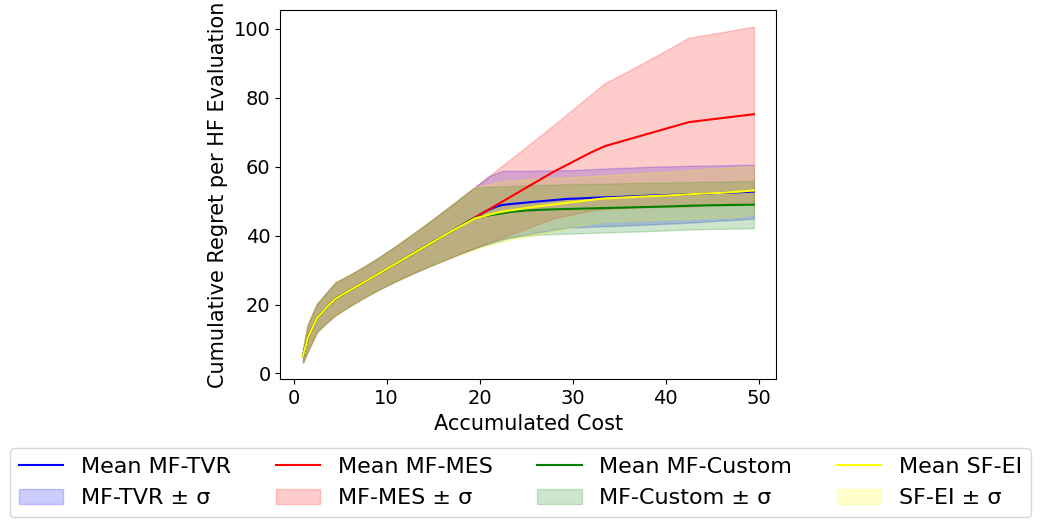}
 \caption{A comparison of the acquisition functions in a single plot for RKHS (Problem 1), using cumulative regret per high-fidelity evaluation. Each experiment was seeded with 5 initial samples (both high- and low-fidelity evaluations), afforded a budget of 50, and had a domain-size of 500. The results are averaged over 5 runs for each search-algorithm.}
 \label{fig:rkhs_new_metric}
\end{figure}

\begin{figure}[h]
    \centering
    \includegraphics[width=1\textwidth]{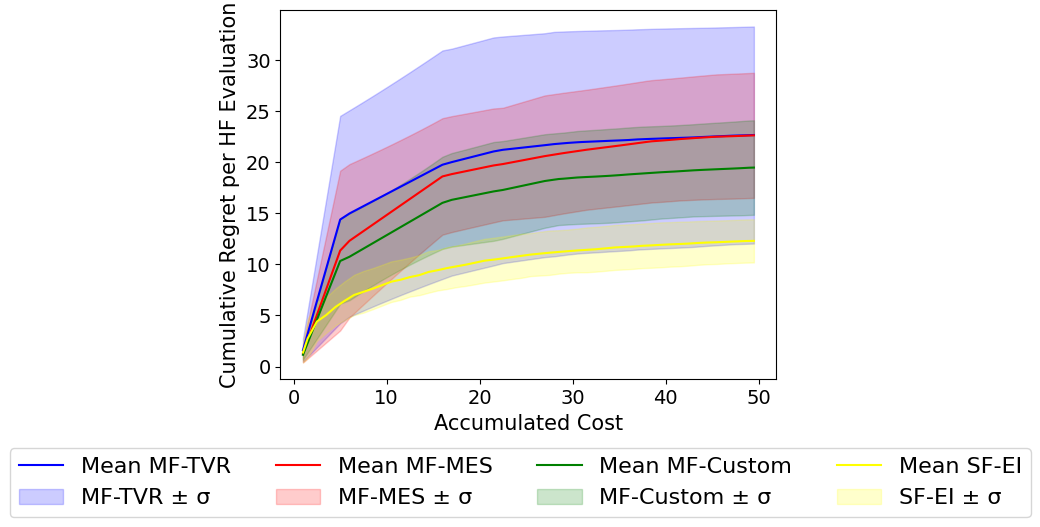}
 \caption{A comparison of the acquisition functions in a single plot for Problem 4, using cumulative regret per high-fidelity evaluation. Each experiment was seeded with 25 initial samples (both high- and low-fidelity evaluations), afforded a budget of 50, and had a domain-size of 44928. The results are averaged over 5 runs for each search-algorithm.} 
 \label{fig:stk_new_metric}

\end{figure}

\end{document}